\pdfoutput=1

\documentclass[11pt]{article}

\usepackage[]{ACL2023}

\usepackage{times}
\usepackage{latexsym}

\usepackage[T1]{fontenc}

\usepackage[utf8]{inputenc}

\usepackage{microtype}

\usepackage{inconsolata}
\usepackage{multirow}
\usepackage{graphicx}
\usepackage{amsmath}
\usepackage{amsthm}
\usepackage{amssymb}
\usepackage{tabularx}
\usepackage{algorithm}
\usepackage{algorithmic}
\usepackage{graphicx}
\usepackage{booktabs} 
\usepackage{enumitem}

\title{Improving Few-Shot Performance of Language Models via Nearest Neighbor Calibration}

\author{
    Feng Nie$^{\ddag}$,\ Meixi Chen$^{\S}$\thanks{\ \ This work was done during the second author's internship at Tencent.},\ Zhirui Zhang$^{\sharp}$\thanks{\ \ Corresponding author.},\ Xu Cheng$^{\ddag}$ \\
    $^{\ddag}$Interactive Entertainment Group, Tencent Inc, Shenzhen, China  \\ $^{\S}$University of Cambridge \ \ $^{\sharp}$Tencent AI Lab \\
    $^{\ddag}$\texttt{\{jannie,alexcheng\}@tencent.com} \ \ $^{\S}$\texttt{mc2125@cam.ac.uk} \ \ $^{\sharp}$\texttt{zrustc11@gmail.com}
}

\begin{document}
\maketitle
\begin{abstract}

Pre-trained language models (PLMs) have exhibited remarkable few-shot learning capabilities when provided a few examples in a natural language prompt as demonstrations of test instances, i.e., \textit{in-context learning}.
However, the performance of in-context learning is susceptible to the choice of prompt format, training examples and the ordering of the training examples.
In this paper, we propose a novel nearest-neighbor calibration framework for in-context learning to ease this issue.
It is inspired by a phenomenon that the in-context learning paradigm produces incorrect labels when inferring training instances, which provides a useful supervised signal to calibrate predictions.
Thus, our method directly augments the predictions with a $k$-nearest-neighbor ($k$NN) classifier over a datastore of cached few-shot instance representations obtained by PLMs and their corresponding labels.
Then adaptive neighbor selection and feature regularization modules are introduced to make full use of a few support instances to reduce the $k$NN retrieval noise.
Experiments on various few-shot text classification tasks demonstrate that our method significantly improves in-context learning, while even achieving comparable performance with state-of-the-art tuning-based approaches in some sentiment analysis tasks.

\end{abstract}

\section{Introduction}

Large-scale pre-trained language models (PLMs), such as BERT and GPT~\cite{bert,gpt18,Radford2019}, have been proven to be fundamental for solving a variety of NLP tasks.
Recently, \citet{gpt3} demonstrate that PLMs can perform few-shot learning when provided a few training examples in a natural language prompt as demonstrations with input sentences, i.e., \textit{in-context learning}.
Specifically, in the sentiment classification task, we use the template ``\texttt{<TEXT>} It was \texttt{[MASK]}.'' for model prediction, where \texttt{<TEXT>} is the placeholder for the input text and the PLMs are asked to infer verbalizers (e.g., `great' and `terrible') for the \texttt{[MASK]} token to score the target labels (e.g., `positive' or `negative'). 
Then each input is further prepended with demonstrations of different sentiments as: ``Formulaic, but fun. It was great. \texttt{[SEP]} Shyamalan should stop trying to please his mom. It was terrible. \texttt{[SEP]} \texttt{<TEXT>} It was \texttt{[MASK]}.''.
This style of few-shot learning is appealing because it shows that the model can directly leverage information from few-shot support instances without parameter updates. 

\begin{figure}[t]
	\centering
    \small
	\includegraphics[width=0.48\textwidth]{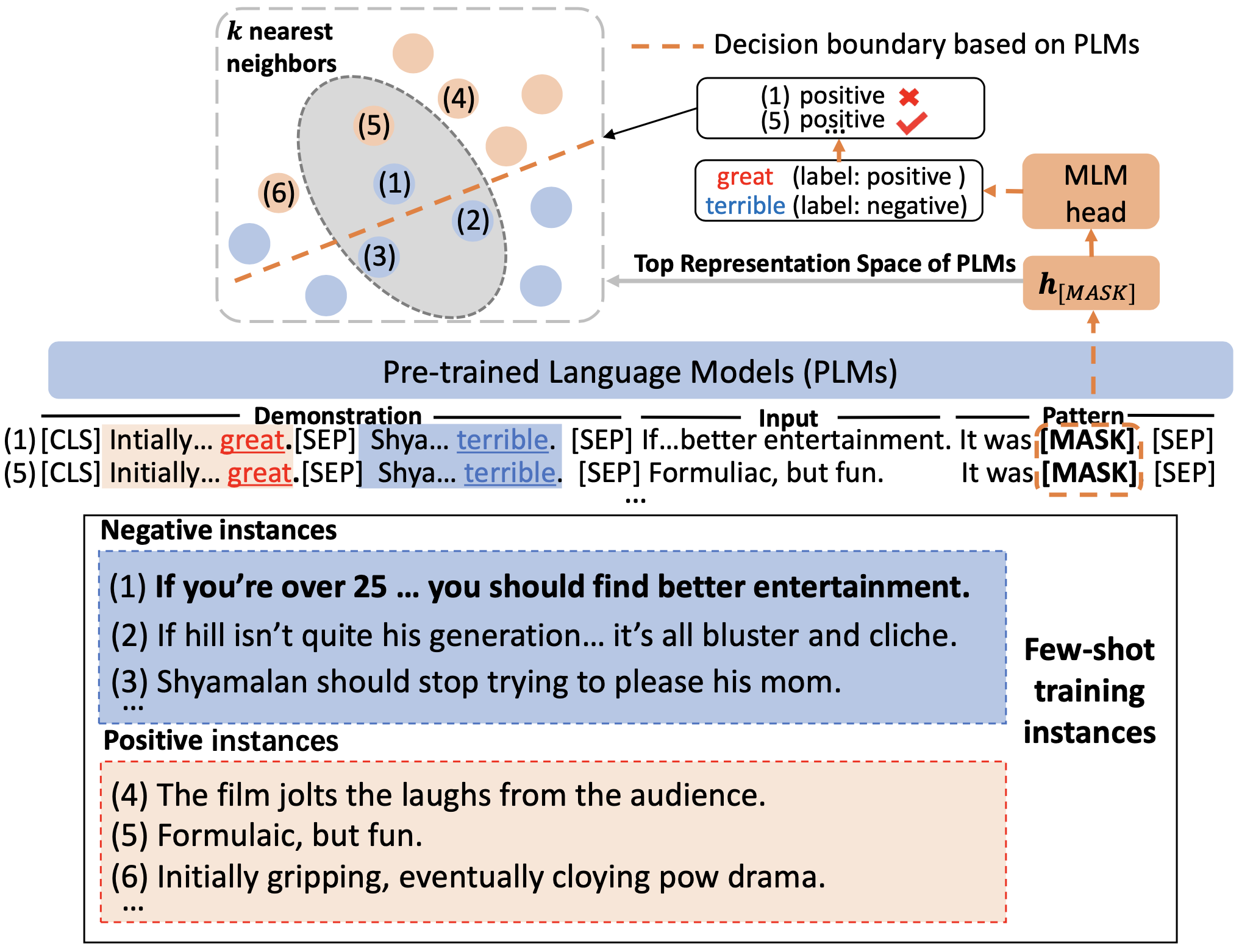}
	\vspace{-12pt}
	\caption{The issue illustration when inferring the label of few-shot training instances themselves via in-context learning. The predictions of in-context learning could conflict with their labels, which indicates the wrong decision boundary provided by PLMs.}
	\label{fig:intro}
\end{figure}

Despite promising results and potential benefits, the in-context learning is highly sensitive to the choice of prompting templates and verbalizers, training examples, and even a permutation (ordering) for training examples, causing accuracy to vary from near chance to near state-of-the-art.~\cite{lester-etal-2021-power,lu-etal-2022-fantastically}.
Another interesting phenomenon is that this approach produces incorrect results when inferring few-shot support instances via in-context learning. 
As illustrated in Figure~\ref{fig:intro}, given an input text with negative label in the training set (e.g., ``If you're over 25 ... you should find better entertainment.''), the remaining support instances are converted into demonstrations but PLMs make a contradicting prediction (e.g., ``positive'') on this input.
The level of contradiction varies with different engineered templates and verbalizers.
This inconsistency actually provides helpful supervised signals and another orthogonal perspective to improve in-context learning.
That is to fully exploit the few-shot learning of PLMs by \textit{vertically} retrieving small similar support samples to correct the decision boundary of PLMs, rather than \textit{horizontally} concatenating them with input texts only. 

In this paper, we propose a simple and effective nearest-neighbor calibration framework to improve the performance of in-context learning in few-shot text classification tasks. 
The whole approach is built on the top hidden representations of PLMs, and then directly enhances PLMs with a $k$-nearest-neighbor ($k$NN) classifier over a datastore of cached few-shot instance representations and their corresponding labels.
In this way, similar training samples of input texts are dynamically retrieved to strengthen or rectify the original prediction distribution provided by PLMs, achieving better model performance and easing the large variance brought by different engineered prompts.
Moreover, the performance of this method largely relies on the quality of $k$NN retrieval, but it may include noise when the inappropriate number of neighbors is adopted or representations produced by PLMs are problematic. 
We further design adaptive neighbor selection and feature regularization modules to reduce the $k$NN retrieval noises with the supervision of current few-shot instances:
the former module is designed to dynamically decide the number of few-shot instances in the $k$NN classifier, while the latter module leverages a lightweight network to separate these instances with different labels but similar representations.

We present a systematic evaluation for analyzing few-shot performance on 6 single-sentence and 6 sentence-pair NLP tasks. 
We observe that given a small number of training examples, (1) our method proves the effectiveness of introducing instance-augmented classification, as it significantly outperforms in-context learning, especially superior in single-sentence tasks with 12.8\% absolute accuracy improvement on average; (2) the proposed method is able to close the performance gap with tuning based methods in single-sentence tasks.

\section{Background}
\paragraph{Task Formulation.}
We consider the few-shot adaption of a pre-trained language model $\mathcal{L}$ on the task $\mathcal{D}$ with a label space $\mathcal{Y}$. 
For the task, we assume that the training data $\mathcal{D}_{\mathrm{train}} = \{(x^i, y^i)\}_{i=1}^{K \times |\mathcal{Y}|}$ only consists of $K$ examples per class, where $x$ represents the input, $y$ is the target label and $|\mathcal{Y}|$ denotes the number of unique classes.
The goal of few-shot adaption is to develop task-agnostic learning strategies on $\mathcal{D}_{\mathrm{train}}$, and generalize well to an unseen test set $\mathcal{D}_{\mathrm{test}}$. 
We additionally assume access to development set $\mathcal{D}_{\mathrm{dev}}$ with the same size as the training data for model selection and hyper-parameter tuning, as larger validation sets can grant a substantial advantage~\cite{perez2021true}.
For our experiments, we use 16 training examples ($K$ = 16) and a development set with 16 examples per class for all tasks.


\begin{figure*}[t]
	\centering
    \small
	\includegraphics[width=0.96\textwidth]{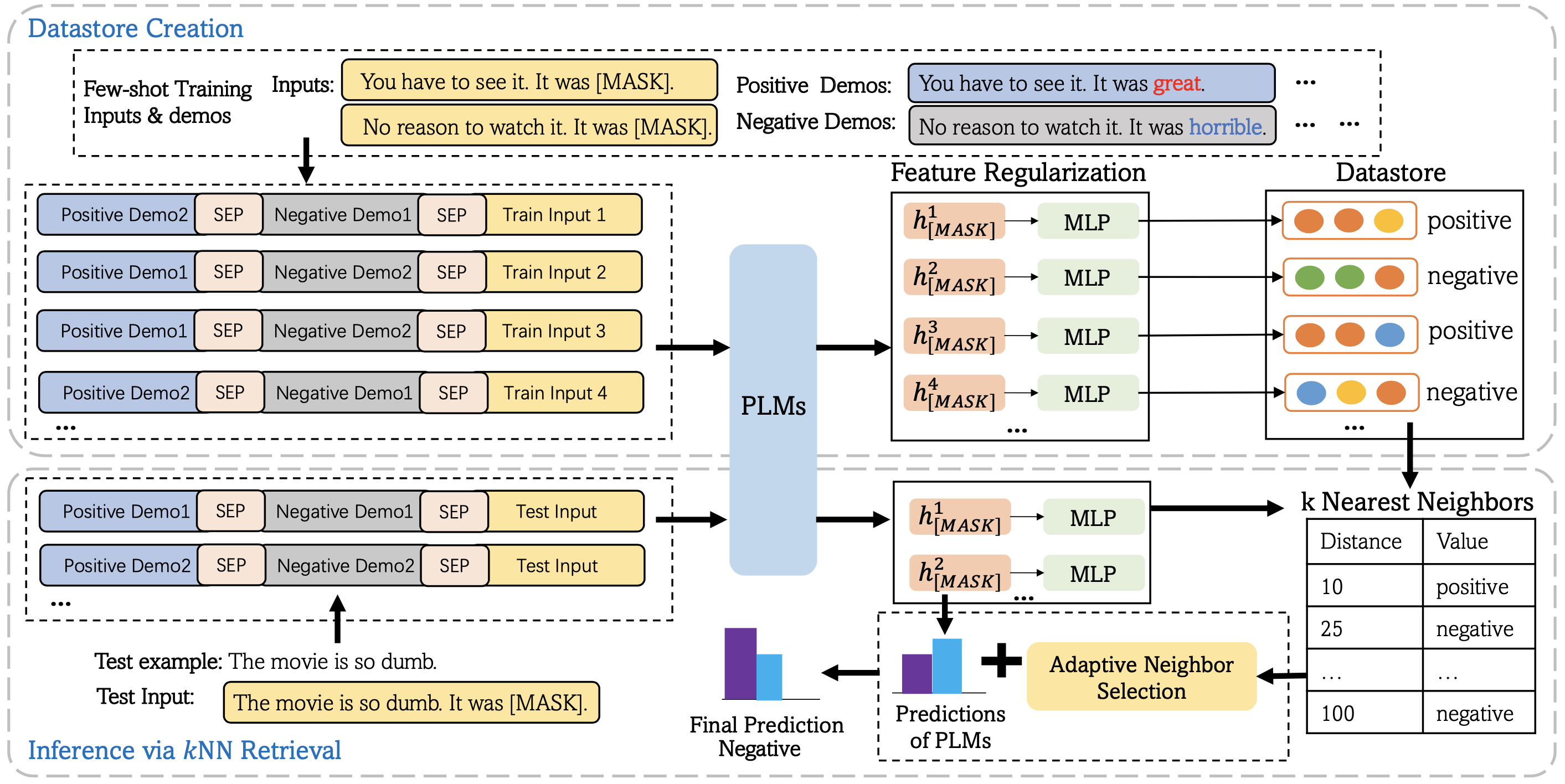}
    \vspace{-8pt}
	\caption{An overview of our nearest neighbor calibration framework (\textbf{KNN-C}) for the sentiment analysis task.} 
	\label{fig:model}
\end{figure*}

\paragraph{Prompt-Based Fine-tuning.}
The standard fine-tuning has a clear discrepancy between pre-training and fine-tuning phases, where the former is optimized by the prediction of masked tokens in the masked language modelling (MLM) task. 
An alternative way to eliminate this gap is prompt-based fine-tuning, in which the task is formulated in a cloze-format~\cite{taylor}. 
In this way, the language model $\mathcal{L}$ predicts label words with the MLM objective. 
Specifically, inputs are converted using a pre-defined prompt template $x_{\mathrm{prompt}} = \mathcal{T}(x)$, e.g., in the sentiment classification task, $x_{\mathrm{prompt}}$ is constructed as follow: 
\begin{align}
    x_{\mathrm{prompt}}=\texttt{[CLS]}\ x \texttt{ It was } \texttt{[MASK]}.\ \texttt{[SEP]}.
    \label{eq:prompt}
\end{align}
Then the language model $\mathcal{L}$ decides which verbalizer (e.g., `great' and `terrible') is most likely for \texttt{[MASK]} in $x_{\mathrm{prompt}}$. 
For the model training, let $\mathcal{M}: \mathcal{Y} \to \mathcal{V}$ be a mapping from the task label space to words in the vocabulary $\mathcal{V}$ of PLMs.
The probability of class $y \in \mathcal{Y}$ is calculated as:
\begin{align}
p(y|x) & = p_{\mathcal{L}}(\texttt{[MASK]}=\mathcal{M}(y)|\mathcal{T}(x)) \\
& = \frac{\mathrm{exp}(\textbf{W}_{\mathcal{M}(y)} h_{\texttt{[MASK]}})}{\sum_{\hat{y} \in \mathcal{Y}}\nonumber  \mathrm{exp}(\textbf{W}_{\mathcal{M}(\hat{y}) }h_{\texttt{[MASK]}})},
\label{promptscore}
\end{align}
where $h_{\texttt{[MASK]}}$ denotes the hidden vector of \texttt{[MASK]} token and  $\textbf{W}_{\mathcal{M}(y)}$ refers to pre-softmax vector for word $v \in \mathcal{V}$. 
The entire model is trained by minimizing the cross-entropy loss with $\mathcal{D}_{\mathrm{train}}$ and select the best checkpoint on $\mathcal{D}_{\mathrm{dev}}$.

\paragraph{In-Context Learning.}
Instead of directly fine-tuning model, \citet{gpt3} show that PLMs themselves have the capability to perform few-shot learning without parameter updates. 
It explores an in-context learning paradigm, which simply concatenates randomly sampled training examples as demonstrations with inputs during inference:
\begin{equation}
\begin{aligned}
x_{\mathrm{demo}} & = \mathcal{T}(x_{train}^{1}) \oplus \ldots \oplus \mathcal{T}(x_{train}^{|\mathcal{Y}|}) \oplus \mathcal{T}(x), \\
p(y|x) & = p_{\mathcal{L}}(\texttt{[MASK]} = \mathcal{M}(y)|x_{\mathrm{demo}}), 
\end{aligned}
\label{equ:demotemplate}
\end{equation}
where $\oplus$ refers to concatenation of input texts, and we select one example per class as demonstrations ($x_{\mathrm{train}}^{1}, ..., x_{\mathrm{train}}^{|\mathcal{Y}|}$).
The final prediction of the in-context learning ensembles all results based on different sampled demonstrations. 
This method has practical advantages for the few-shot adaptation over the now-standard approach of finetuning, as we could hold only one model for serving many different tasks, avoiding expensive and time-consuming parameter updates.    

\section{Nearest Neighbor Calibration}
In this paper, we attempt to further explore the potential of PLMs for few-shot adaptation along the research direction of in-context learning.
As shown in Figure~\ref{fig:intro}, inference over few-shot training instances via in-context learning may produce contradictory results, and the situation can be exaggerated by poor templates and verbalizers.
This phenomenon motivates us to exploit few-shot support instances by retrieving similar training samples to strengthen or rectify the original prediction distribution generated by PLMs. 
In this work, we propose a novel nearest-neighbor calibration framework for in-context learning, as illustrated in Figure~\ref{fig:model}.
Our method is built on the top representations of PLMs and it augments classification with a retrieval pipeline, which directly queries a datastore of cached few-shot instance representations and corresponding labels to produce prediction distribution. We further design lightweight adaptive neighbor selection and feature regularization modules to reduce retrieval noises. Next, we first introduce the instance-augmented classification, and then detail the adaptive neighbor selection and feature regularization modules.

\subsection{Instance-Augmented Classification}
\label{sec:ds}

We achieve this instance-augmented classification through a $k$NN classifier, which consists of two steps: creating a datastore and making predictions depending on it.
\paragraph{Datastore Creation.} 
The datastore is the cache of a set of key-value pairs, which is constructed by the training set $\mathcal{D}_{\mathrm{train}} = \{(x^i, y^i)\}_{i=1}^{K \times |\mathcal{Y}|}$ using language model $\mathcal{L}$. 
Firstly, we convert input $x$ through a prompt template, and then concatenate it with demonstrations sampled from the remaining few-shot training examples following Eq.~(\ref{equ:demotemplate}), yielding $x_{\mathrm{demo}}$. 
We further feed $x_{\mathrm{demo}}$ into language model $\mathcal{L}$ to obtain the hidden representation of \texttt{[MASK]} token (i.e., the representation of last transformer layer), denoted as $f(x_{\mathrm{demo}}; \mathcal{L}) = h_{\texttt{[MASK]}}$. 
Thus, the whole datastore is constructed by taking $h_{\texttt{[MASK]}}$ as key and corresponding label $y$ as value:
\begin{align}
\bigcup_{(x, y) \in \mathcal{D}_{\mathrm{train}}} \{(f(x_{\mathrm{demo}}; \mathcal{L}), y), \forall x_{\mathrm{demo}} \in \mathcal{S}(x) \},
\notag
\end{align}
where $\mathcal{S}(x)$ is the set of $x$ with different demonstrations. 
We randomly sample $K$ times without replacement to involve all training samples, resulting in $K \times K \times |\mathcal{Y}|$ records of datastore in total.

\paragraph{Inference via $k$NN Retrieval.}
After building the datastore, we augment the prediction of PLMs with $k$NN retrieval similar to \citet{knnmt}. 
Specifically, the test instance $x^t \in \mathcal{D}_{\mathrm{test}}$ is converted into $x^t_{\mathrm{demo}}$ following the same data pre-processing process as the datastore construction, and we obtain corresponding vector representation $f(x^t_{\mathrm{demo}}; \mathcal{L})$ of \texttt{[MASK]} token for subsequent $k$NN retrieval.
Then the $k$NN-based prediction distribution $p_{k\mathrm{NN}}$ over the label space  $\mathcal{Y}$ is obtained with nearest neighbors:
\begin{equation}
\begin{aligned}
    p_{k\mathrm{NN}} (y|x^t_{\mathrm{demo}}) & \propto \\
     \sum_{(h_i,v_i) \in N_t} \mathrm{I}_{y=v_i} & \cdot \mathrm{exp} (\frac{-d(h_i, f(x^t_{\mathrm{demo}}; \mathcal{L}))^2}{\tau}),
\end{aligned}
\label{equ:knn-prob}
\end{equation}
where $d(.,.)$ stands for euclidean distance, $N_t$ represents the set of $k$ nearest neighbours, and $\tau$ is the temperature to control the sharpness of softmax function.
The final prediction distribution of $y$ is calculated as the interpolation of two distributions with a tuned hyper-parameter $\lambda \in [0,1]$:
\begin{equation}
\begin{aligned}
     p(y|x^t) & = (1 - \lambda) \cdot p_{k\mathrm{NN}}(y|x^t_{\mathrm{demo}}) \\
      & + \lambda \cdot p_{\mathcal{L}}(\texttt{[MASK]} = \mathcal{M}(y)|x^t_{\mathrm{demo}}).
\end{aligned}
\label{equ:knn-pred}
\end{equation}
Note that we use the development set to select the appropriate hyper-parameters (interpolation factor $\lambda$, number of nearest neighbor $k$ and temperature $\tau$). 
Similar to in-context learning, we also ensemble all results with different demonstrations for each test instance during inference. 

\subsection{Improving Robustness of $k$NN Retrieval}
The performance of instance-augmented classification highly relies on the quality of $k$NN retrieval.
However, the retrieved nearest neighbors typically include noises due to the inappropriate pre-defined neighbor selection and messy vector space produced by PLMs.
To address these issues, we design the adaptive neighbor selection and feature regularization modules, which utilize current all few-shot training instances to mitigate retrieval noise as much as possible.

\paragraph{Adaptive Neighbor Selection (ANS).}
Following \citet{zheng-etal-2021-adaptive}, instead of a pre-defined $k$, we consider a set of possible $k$s smaller than an upper bound $k_{\mathrm{max}}$ and introduce a lightweight network for the importance estimation of utilizing different selections to reduce the risk of inappropriate neighbors.
In practice, we consider multiples of $4$ as the choices of $k$ for simplicity, alone with $k = 0$ utilizing only PLMs, i.e., $k \in \mathcal{A}$ where $\mathcal{A}=\{0 \} \cup \{ k_i \in \mathbb{N} \mid k_i/4 \in \mathbb{N}, k_i \leq k_{\mathrm{max}} \}$. 
Then the lightweight network evaluates the confidence of different $k$NN retrieval results by taking retrieved neighbors as inputs. 

Concretely, for test instance with a demonstration $x^t_{\mathrm{demo}}$, we first retrieve $k_{\mathrm{max}}$ neighbors $N_{t}$ from the datastore and compute their distance from the current representation, as well as the count of distinct values in top $i$ neighbors $c_i$.
We take the computed distances $d=(d_1, ..., d_{k_{\mathrm{max}}})$ and counts $c = (c_1, ..., c_{k_{\mathrm{max}}})$ of distinct values in corresponding labels $v=(v_1, ..., v_{k_{\mathrm{max}}})$ as model inputs to decide the best $k$.
Intuitively, the distance of each neighbor is the most direct evidence when evaluating their importance. 
The intuition to take label counts $c$ as model input is that inference should trust more on PLMs when retrieved labels are chaotic.
The importance estimation network $f_{\textrm{ANS}}(\cdot)$ is the two feed-forward layers with a non-linear function between them, in which the hidden size is set to $32$ and we adopt $g(.) = \mathrm{ReLU}(.)$~\cite{relu} as the non-linear function.
The probability of selecting $k$ is calculated as follows:
 \begin{align}
     p_{a}(k|x^t_{\mathrm{demo}}) = \mathrm{softmax}(f_{\textrm{ANS}}([d, c])).
\end{align}
Instead of introducing the hyper-parameter $\lambda$ as Eq.~(\ref{equ:knn-pred}),
we aggregate the output of PLMs and different $k$NN predictions with the importance estimation network for final prediction:
\begin{equation}
p(y|x^t) = \sum_{k_i \in \mathcal{A}} p_{a}(k_i|x^t_{\mathrm{demo}}) \cdot p_{k_i\textrm{NN}}(y|x^t_{\mathrm{demo}}),
\label{equ:adaptive-prob}
\end{equation}
where $p_{k_i\textrm{NN}}$ indicates the $k_i$-nearest-neighbor prediction results calculated as Eq.~(\ref{equ:knn-prob}).
For training this lightweight network, we randomly split the $\mathcal{D}_{\mathrm{train}}$ into two equal parts.
One part is used to build the datastore, while we train $f_{\textrm{ANS}}(\cdot)$ on another part via minimizing the cross-entropy loss following Eq.~(\ref{equ:adaptive-prob}).

\paragraph{Feature Regularization (FR).}
When facing the situation of instances having different labels but sharing similar representations, the $k$NN classifier easily fails or makes mistakes, since retrieval results are noisy and the ANS module tends to ignore this case.
To mitigate this, we further leverage a simple linear layer to separate such instance representations using the supervision of all few-shot instances. 
For key-value pair $(h_i, v_i)$ in the datastore, we reconstruct the representation as follows:
\begin{align}
    h_i^{f} = g(\textbf{W}_{f}h_i + b_{f}),
    \label{eq:ftlayer}
\end{align}
where $\textbf{W}_{f} \in \mathbb{R}^{H \times Z}$ and $b_{f} \in \mathbb{R}^{Z}$ are trainable parameters, and $Z$ is the new dimension of representation space.
As the training data is very small, we select a small $Z$ to avoid overfitting.
We set $Z=32$ for our experiments to reduce the number of trainable parameters and the new datastore is constructed as $\{(h_i^{f}, v_i)\}$.
This FR module targets to make representations belonging to the same class semantically close by maximizing $k$NN retrieval probability in Eq.~(\ref{equ:knn-prob}).
Similar to the training process of the ANS module, we optimize this linear layer by minimizing the cross-entropy loss of $k$NN classifier. 
As for combining ANS and FR modules, we first train the FR module with two equal parts of $\mathcal{D}_{\mathrm{train}}$, and then exchange these two parts for the optimization of ANS module.

\section{Experiments}
We conduct extensive experiments on a wide variety of NLP tasks, and compare the performance of our proposed approach with existing the-state-of-art methods in the few-shot scenario.

\subsection{Experimental Setup}
\paragraph{Datasets.} 
We evaluate methods on 12 datasets for 7 tasks:
(1) sentiment analysis datasets, SST-2~\cite{sst}, SST-5~\cite{sst}, CR~\cite{hu2004mining} and MR~\cite{mr};
(2) the subjectivity classification dataset, SUBJ~\cite{subj};
(3) the question classification dataset, TREC~\cite{trec};
(4) natural language inference datasets, CB~\cite{de2019commitmentbank} and RTE~\cite{rte};
(5) the question answering dataset, QNLI~\cite{qnli};
(6) paraphrase detection datasets, MRPC~\cite{mrpc} and QQP\footnote{https://quoradata.quora.com};
(7) the word sense disambiguation dataset, WiC~\cite{wic}.
Following \citet{karimi-mahabadi-etal-2022-prompt}, we evaluate on original test sets for MR, CR, SST-5, SUBJ, and TREC. For the remaining datasets, we test on original validation set. 
We sample $K=16$ examples per class from original training data to form few-shot training and validation sets.

\paragraph{Baselines.}
We adopt RoBERTa-large~\cite{roberta} as the underlying languge model $\mathcal{L}$ for all methods in our experiments, and compare our approach (\textbf{KNN-C}) with the following baselines:
(i) \textbf{PV-Zero}: we take manual prompts and verbalizers to obtain prediction of PLMs without involving any training examples;
(ii) \textbf{In-Context Learning (ICL)}~\cite{gpt3}: we adopt the same prompt formats as in PV-Zero, but augment the context with randomly sampled demonstrations (and still use RoBERTa-large, not GPT-3). Note that we sample $K$ demonstrations from the training set without replacement to cover all training samples during inference;
(iii) \textbf{Contextual Calibration (ICL+CC)}: following \citet{calibrate}, we introduce content-free input ``N/A'' to remove the prediction bias of in-context learning;
(iv) \textbf{Fine-tuning (FT)}: the standard fine-tuning method~\cite{bert}, introducing a classifier on top of the \texttt{[CLS]} token and tuning all parameters of PLMs;
(v) \textbf{PET}~\cite{pet}: the prompt-based fine-tuning method that employs manual prompts and requires fine-tuning over PLMs;
(vi) \textbf{PERFECT}~\cite{karimi-mahabadi-etal-2022-prompt}: the-state-of-art fine-tuning method for few-shot adaptation, which avoids manual prompts and verbalizers by introducing adapters and multi-token label embeddings. 

\begin{table*}[t]
\small
\centering
\scalebox{0.91}{
\begin{tabular}{l|cccccc|c}
\hline  
\toprule
\multicolumn{1}{l|}{Methods} & \multicolumn{1}{c}{\textbf{SST-2}} & \multicolumn{1}{c}{\textbf{SST-5}} & \multicolumn{1}{c}{\textbf{CR}}   & \multicolumn{1}{c}{\textbf{MR}}   & \multicolumn{1}{c}{\textbf{SUBJ}} & \multicolumn{1}{c}{\textbf{TREC}} & \multicolumn{1}{|c}{\textbf{AVG}} \\ \midrule
PV-Zero$^{*}$   & 63.4/54.0/11.8 & 27.9/21.1/4.8   & 59.6/50.4/11.8  & 62.6/53.7/10.9                    & 55.5/50.2/7.6   & 33.9/25.6/8.8     & 50.5/42.5/9.3    \\
ICL & 89.9/\textbf{88.3/1.0}  & 43.6/37.3/4.8  & 88.0/83.1/2.5 & 86.1/\textbf{83.9/1.2}  & 51.2/48.5/2.9   &  52.7/36.8/10.3   & 68.6/63.0/3.8  \\ 
ICL+CC  & 84.6/82.3/1.9  & 38.3/33.2/3.8 & 85.4/80.9/2.5 & 80.7/77.1/2.3  & 53.4/50.5/2.8   &  49.8/33.4/9.1   & 65.4/59.6/3.7  \\ 
FT    & 81.4/70.0/4.0  & 39.2/34.3/2.5  & 80.1/72.9/4.1  & 77.7/66.8/4.6  & \textbf{90.2/84.1/1.8}    & 87.6/75.8/3.7   & 76.0/67.3/3.4   \\
PET    & 89.7/81.0/2.4   & \textbf{45.9/40.3/2.4}  & 88.4/68.8/3.0   & 85.9/79.0/2.1   & 88.1/79.6/2.4    & 85.0/70.6/4.5   & 80.5/69.9/2.8  \\
PERFECT   & \textbf{90.7}/88.2/1.2  & 42.7/35.1/2.9  & \textbf{90.0/85.5/1.4}  & \textbf{86.3}/81.4/1.6  & 89.1/82.8/2.1  & \textbf{90.6/81.6/3.2}  & \textbf{81.6/75.8/2.1}  \\ \midrule
KNN-C     & 92.6/90.1/0.8  & \textbf{48.5/41.5/2.5}  &  \textbf{90.5}/86.0/\textbf{1.3}   & 89.0/85.8/1.2  & 90.5/86.9/1.5 & \textbf{77.3/64.2/7.9}   & \textbf{81.4/75.7/2.5}  \\ 
- ANS     & 91.9/89.2/1.0  & 46.7/40.8/2.7  &  90.4/84.9/1.7  & 88.3/82.8/1.6  & 90.4/87.5/1.4  & 76.5/64.0/8.2   &  80.7/74.9/2.8 \\
- FR      & \textbf{92.7}/90.6/0.7  & 47.6/40.1/3.1  & 89.9/\textbf{86.9}/1.6           & \textbf{89.5/87.3/0.8}  & \textbf{90.7}/86.3/1.3 & 69.8/46.2/10.2  & 80.0/72.9/3.0 \\
- ANS,FR      & 91.2/88.3/1.5  & 47.1/40.1/3.0  & 89.3/84.9/1.7            & 88.8/85.9/1.3  &  89.9/83.7/2.3 & 69.8/50.0/10.4 & 79.3/72.1/3.4 \\
- ANS,FR,$p_{\mathcal{L}}$  &  92.4/\textbf{90.9/0.7}  & 44.8/38.7/3.2 & 89.6/86.7/1.8           & 88.9/87.1/1.0  & 90.6/\textbf{88.2/1.2} & 65.6/46.4/9.5   & 78.7/73.0/2.9 \\
\midrule
\multicolumn{1}{l|}{Methods} & \multicolumn{1}{c}{\textbf{CB}}    & \multicolumn{1}{c}{\textbf{RTE}}   & \multicolumn{1}{c}{\textbf{QNLI}} & \multicolumn{1}{c}{\textbf{MRPC}} & \multicolumn{1}{c}{\textbf{QQP}}  & \multicolumn{1}{c}{\textbf{WiC}}  & \multicolumn{1}{|c}{\textbf{AVG}} \\ \midrule 
PV-Zero$^{*}$  & 64.3/58.9/3.4   & 56.7/\textbf{54.2}/1.8  & 50.1/49.4/\textbf{0.6}  & 67.5/\textbf{66.4/0.7} & 36.8/36.8/\textbf{0.1}   & 51.4/48.8/2.9   & 54.5/52.4/\textbf{1.6}  \\
ICL & 77.4/57.1/10.3 & 58.8/53.8/3.6 & 52.0/50.3/1.2 & 54.0/32.1/17.4  & 42.5/36.8/6.7   & 51.0/\textbf{50.0}/1.6  & 55.9/46.7/6.8 \\
ICL+CC & 56.9/42.9/10.0 & 56.0/53.8/\textbf{1.6} & 54.4/52.8/0.8 & 62.2/37.8/10.4  & 45.0/36.7/6.6   & 49.3/47.5/\textbf{1.0} & 54.0/45.2/5.1  \\ 
FT  & 72.9/67.9/\textbf{2.5}  & 56.8/50.2/3.5 & 62.7/51.4/7.0   & \textbf{70.1}/62.7/4.7 & 65.0/59.8/3.6   & 52.4/46.1/3.7   & 63.3/56.4/4.2  \\
PET     & 86.9/73.2/5.1   & 60.1/49.5/4.7   & 66.5/55.7/6.2  & 62.1/38.2/6.8 & 63.4/44.7/7.9 & 51.0/46.1/2.6   & 65.0/51.2/5.6 \\
PERFECT  & \textbf{90.3/83.9}/3.5  & \textbf{60.4}/53.1/4.7 & \textbf{74.1/60.3}/4.6  & 67.8/54.7/5.7  & \textbf{71.2/64.2}/3.5 & \textbf{53.8}/47.0/3.0 & \textbf{69.6/60.5}/4.2  \\ \midrule
KNN-C     & \textbf{82.1}/69.6/5.3  & \textbf{61.8}/52.4/\textbf{2.9} & \textbf{54.2}/50.4/2.2    & 62.5/42.7/7.3 & 58.7/47.4/\textbf{4.3} & 53.1/46.2/3.1  & \textbf{62.1}/51.4/\textbf{4.2} \\ 
- ANS     & 79.8/64.3/6.1 & 60.7/51.3/3.1 & 54.0/50.2/2.2   & 62.1/47.1/\textbf{6.0}  & \textbf{59.6/47.8}/4.6 & 52.8/45.6/2.9  & 61.5/51.0/\textbf{4.2} \\
- FR      & 80.9/\textbf{71.4/5.1}  & 60.9/51.3/3.7  & 52.9/50.9/1.2  & 63.8/41.4/7.4  & 54.5/45.3/5.0  & \textbf{55.0/50.2}/2.8   & 61.3/\textbf{51.7/4.2} \\
- ANS,FR      & 80.9/69.6/\textbf{5.1}  & 61.0/\textbf{53.8}/3.6  & 53.1/\textbf{51.4/1.1}   & 61.4/37.8/9.3  & 53.2/38.1/5.1  & 53.4/50.0/\textbf{2.5}  & 60.5/50.1/4.4  \\
- ANS,FR,$p_{\mathcal{L}}$  & 77.6/67.9/6.3  & 57.4/49.5/4.1  & 52.2/50.3/\textbf{1.1} & \textbf{63.9/49.0}/6.7  & 56.1/47.4/4.5 & 55.1/50.8/2.6 & 60.4/52.5/\textbf{4.2} \\
\bottomrule
\hline
\end{tabular}
}
\vspace{-8pt}
\caption{Performance of all methods on 6 single-sentence (top) and 6 sentence-pair (bottom) benchmarks. $*$: no training examples are used. We report the average accuracy($\uparrow$)/worst-case accuracy($\uparrow$)/standard deviation($\downarrow$) and bold fonts indicate the best results.}
\label{table:main-result}
\end{table*}

\paragraph{Implementation Details.} 
For PLMs, we use the HuggingFace Pytorch implementation. 
In our experiments, we adopt the manually designed patterns and verbalizers used in \citet{karimi-mahabadi-etal-2022-prompt} (usually 5 different options for each dataset).
We evaluate all methods using 5 different random samples for creating the training/development sets and 4 different random seeds for model training.
Thus, we report the results of $5 \times 5=25$ runs for PV-Zero, ICL, and ICL+CC, while we perform $20 \times 5=100$ runs for methods that involve hand-crafted prompts and training process, such as PET and KNN-C.
We perform $5 \times 4 = 20$ runs for methods without hand-crafted prompts, i.e., FT and PERFECT. 
As the variance of each method is usually high in few-shot learning~\cite{perez2021true,calibrate}, we report average accuracy, worst-case performance, and the standard deviation across all runs. 
Note that we adopt the same random samples and seeds as \citet{karimi-mahabadi-etal-2022-prompt} to achieve a fair comparison.
We run all the experiments on one NVIDIA A100 with 40G of memory.
For the training process of KNN-C, we deploy the Adam optimizer~\cite{adam} with a learning rate of 1e-3, batch size and total epoch are set to 64 and 30. 
We use the development set and Eq.~(\ref{equ:knn-pred}) to select best hyper-parameters: interpolation factor $\lambda$, temperature $\tau$ and the number of retrieved neighbors $k$.
We find that $\tau$ and $k$ are less sensitive to downstream tasks, thus $\tau=5$ and $k=8$ are used for all tasks and we set $k_{\mathrm{max}}$ to 16 for our experiments;

\subsection{Main Results}
Table~\ref{table:main-result} lists the performance of different methods on 6 single-sentence (top) and 6 sentence-pair (bottom) benchmarks. Baselines are divided into two categories: ICL-based and FT-based methods.

\paragraph{KNN-C vs. ICL-based Methods (PV-Zero, ICL, and ICL+CC).}
PV-Zero is almost the worst except for the MRPC dataset, indicating an obvious discrepancy between pre-training and downstream tasks.    
ICL introduces a few training samples with the format of demonstrations and generally performs better than PV-Zero. 
It should be noted that improvements of ICL are much smaller in pairwise tasks compared to those in single sentence tasks (1.4\% vs. 18.1\% on average accuracy), indicating that transferring knowledge to pairwise tasks (e.g., text entailment) is \textit{a lot more difficult} than to single-sentence tasks (e.g., sentiment analysis).
We argue this is because PLMs take inputs sentence by sentence, rather than sentence pairs during the pre-training, leading to poor few-shot performance of in-context learning.
We also observe that the contextual calibration method hurts the few-shot performance of the Roberta-large model in most datasets.
Compared with ICL, KNN-C further refines the decision boundary generated by PLMs with the help of similar retrieval instances from training data. 
It achieves $12.8\%$ and $6.2\%$ absolute improvements on average accuracy for single-sentence and sentence-pair datasets, respectively.

\begin{table*}[t]
\centering
\small
\scalebox{0.95}{
\begin{tabular}{l|c|c|ccccc}
\hline    
\toprule
\multicolumn{1}{c|}{\multirow{2}{*}{Dataset}} & \multicolumn{1}{c|}{\multirow{2}{*}{PV-Zero}} & \multicolumn{1}{c|}{\multirow{2}{*}{ICL}} & \multicolumn{5}{c}{KNN-C}                                     \\ 
\multicolumn{1}{c|}{}       & \multicolumn{1}{c|}{}    & \multicolumn{1}{c|}{}                                & w/o Aug        & w/ Aug(1)     & w/ Aug(4) & w/ Aug(8)     & w/ Aug(Full) \\ \midrule
SST-2                                                  & 63.4/54.0/11.8   &   89.9/88.3/1.0                                   & 83.4/63.7/5.3  & 90.0/83.3/1.7 & 91.7/87.8/1.2    & 92.3/90.0/0.9 & \textbf{92.6/90.1/0.8}      \\
SST-5                                                  & 27.9/21.1/4.8    &  43.6/37.3/4.8                                      & 37.9/29.1/3.4  &  44.9/38.6/2.7 & 47.8/41.5/\textbf{2.4}    & 48.4/\textbf{42.2/2.4} & \textbf{48.5}/41.5/2.5    \\
CR                                                     & 59.6/50.4/11.8   & 88.0/83.1/2.5                                     & 80.3/73.3/3.2  & 87.7/83.3/2.0 & 89.4/\textbf{86.1}/1.4 & 90.2/85.6/\textbf{1.2} & \textbf{90.5}/86.0/1.3       \\
MR                                                     & 62.6/53.7/10.9   & 86.1/83.9/1.2                                    & 80.4/55.9/4.8 & 86.3/77.8/2.0 &  88.6/\textbf{86.2/1.1}  & 88.8/85.7/1.3 &  \textbf{89.0}/85.8/1.2     \\
SUBJ                                                   & 55.5/50.2/7.6   &   51.2/48.5/2.9                                     & 86.6/77.7/2.8 & 88.9/81.2/2.1 &  89.8/84.8/1.9 & 90.3/86.0/1.7 & \textbf{90.5/86.9/1.5}  \\
TREC                                                   & 33.9/25.6/8.8   &   52.7/36.8/10.3                                   & 66.9/28.0/18.0 & 76.5/62.6/\textbf{7.2} & 76.8/63.2/8.1 & 77.1/63.8/8.0 & \textbf{77.3/64.2}/7.9 \\ \midrule
CB                                                     & 64.3/58.9/3.4   &   77.4/57.1/10.3                                      & 68.6/57.1/4.3 & 72.5/39.3/12.7 & 80.4/60.7/7.0  & 81.7/71.4/5.5  &  \textbf{82.1/69.6/5.3}      \\
RTE                                                    & 56.7/54.2/1.8   &  58.8/\textbf{53.8}/3.6                                       & 55.9/47.7/3.0  & 61.1/52.7/3.0  &  61.6/53.4/\textbf{2.7} & 61.6/53.1/2.8  &  \textbf{61.8}/52.4/2.9       \\
QNLI                                                   & 50.1/49.4/\textbf{0.6}  &  52.0/50.3/1.2                                    & 52.2/48.6/2.0  & 53.8/50.3/2.2 & 54.1/50.7/2.2  & 54.1/\textbf{50.8}/2.3 & \textbf{54.2}/50.4/2.2     \\
MRPC                                                   & \textbf{68.4/68.1/0.2}  & 54.0/32.1/17.4                                        & 58.7/41.9/6.8 &  62.4/44.9/5.9 &  63.1/41.2/6.7   & 62.8/38.7/6.8 & 62.5/42.7/7.3     \\
QQP                                                    & 36.8/36.8/\textbf{0.1}  &  42.5/36.8/6.7    &  55.7/38.5/7.9  &  56.8/40.4/6.2  & 58.2/44.1/4.9      & 58.6/\textbf{47.4}/4.3 &  \textbf{58.7/47.4}/4.3         \\
WiC                                                    & 51.4/48.8/2.9   & 51.0/\textbf{50.0/1.6}                                      & 51.1/45.0/1.8  & 51.9/44.2/2.7 & 52.8/44.0/2.7  &  53.1/44.7/3.0 & \textbf{53.1}/46.2/3.1  \\ \midrule
AVG                                                    & 52.6/47.6/5.4  &  62.3/54.8/5.3                                       & 64.9/50.5/5.3  &  69.4/58.2/4.2 & 71.2/62.0/3.5     &  71.6/63.3/\textbf{3.4} & \textbf{71.7/63.6/3.4}  \\
\bottomrule
\hline
\end{tabular}
}
\vspace{-8pt}
\caption{The performance comparisons of using $k$NN retrieval and demonstration, including only $k$NN retrieval (KNN-C w/o Aug), only demonstration (ICL), and their combination. We report the average accuracy($\uparrow$)/worst-case accuracy($\uparrow$)/standard deviation($\downarrow$) and bold fonts indicate the best results.}
\label{table:aug}
\end{table*}

\paragraph{KNN-C vs. FT-based Methods (FT, PET and PERFECT).}
Our approach largely closes the performance gap between ICL-based methods and existing state-of-the-art FT-based methods in both single and pairwise sentence tasks.
For single-sentence tasks, KNN-C achieves the best average accuracy performance in 4 out of 6 tasks, and the overall performance is comparable with PERFECT. 
For sentence-pair tasks, transferring knowledge from PLMs to downstream tasks is difficult, and a large performance gap still exists, especially for text entailment and paraphrasing tasks.
Our pilot study also presents the potential of ICL to gain comparable performance with FT-based methods.
We also include the performance comparisons and discussions with more parameter-efficient fine-tuning methods in the Appendix~\ref{sec:appendix-compare-ft}.

\paragraph{Ablation Study.}
We further verify the impact of AFS and FR modules in reducing the noises of $k$NN retrieval.
The ANS module works well on reducing $k$NN retrieval noises in most cases, except for the QQP dataset.
It proves the effectiveness of dynamically controlling the number of similar instances. 
The FR module brings better results and reduces the standard deviation on most datasets, especially for TREC and QQP datasets.
These results show that the FR module helps regularize representations of the same class to be similar, when the gap between pre-training and downstream tasks is relatively large.
But these two modules are not always complementary in our experiments. 
We believe that optimizing these two modules with small data sets results in instability and sometimes fails to search for the optimal solution. 
We also test the performance of our method when removing $p_{\mathcal{L}}$, i.e., we only use $p_{k\mathrm{NN}}$ for prediction.
It outperforms ICL on average accuracy, indicating that errors made by ICL could be remedied by our retrieval strategy. 

\subsection{Analysis}

\paragraph{Effect of Instance-Augmented Retrieval.}
Instead of concatenating training instances with test inputs, our method explores another orthogonal direction that attempts to retrieve similar support samples to augment or correct the original decision boundary.
It can be also applied to PV-Zero, namely ``w/o Aug''.
We compare the performance of these two different ways and verify their combination in different sampling settings.
Table~\ref{table:aug} illustrates the performance comparisons on all tasks, where ``w/ Aug(1/4/8/Full)'' denotes that we augment the context with 1/4/8/$K$ sampled demonstrations. 
We can see that ``w/o Aug'' significantly outperforms PV-Zero, which verifies the effectiveness of our proposed method again. 
ICL and ``w/o Aug'' end in a tie, but shines in different tasks. 
Specifically, ``w/o Aug'' outperforms on SUBJ, TREC and QQP, while ICL works well on SST-5, CR and CB. 
By combining these two methods, KNN-C achieves significant improvements over ``w/o Aug'', even with only one sampled demonstration.
The performance of KNN-C continues to improve with the increase of augmented demonstrations for each instance.
These results prove that our approach is complementary to ICL and could further leverage the potential of PLMs on the few-shot adaptation.

\paragraph{Comparisons of BERT and RoBERTa.}
Table~\ref{table:bert-vs-roberta} compares the performance of using BERT-large (cased) as the PLMs backbone.
According to previous results (in Table~\ref{table:aug}), KNN-C achieves significant improvements over the baselines when using either BERT or RoBERTa as the PLMs backbone.
In addition, the improvement of ICL over PV-Zero on RoBERTa is significantly better than that on BERT, showing that RoBERTa performs better few-shot learning. 
Instead, our method benefits both models and shows better generalization ability.

\paragraph{Improvements vs. ICL Errors on Train Data. }
From Table~\ref{table:icl-errors}, we see that ICL may make wrong predictions on train data, whose performance is similar to one on test data. 
KNN-C tends to achieve bigger improvements on the dataset that ICL performs poorly, as our method leverages $k$NN retrieval to calibrate predictions.

\begin{table}[t]
\centering
\small
\scalebox{0.95}{
\begin{tabular}{l|ccc}
\hline    
\toprule
\multicolumn{1}{c|}{\multirow{2}{*}{Dataset}} & \multicolumn{3}{c}{BERT-large}  \\
\multicolumn{1}{c|}{}       &  PV-Zero & ICL & KNN-C  \\ \midrule
SST-2    & 59.8/52.1/8.1  & 64.3/49.2/9.9 & \textbf{76.6/71.3/3.0}    \\
SST-5    & 26.2/23.6/2.7 & 28.3/24.2/2.6 & \textbf{35.9/30.7/1.9}     \\
CR       & 54.5/50.0/5.0   & 70.6/59.4/6.6 &\textbf{ 77.0/67.5/2.5}    \\
MR       & 58.1/50.8/7.0  & 64.7/51.9/7.7  & \textbf{75.4/69.2/2.4}  \\
SUBJ     & 53.7/50.2/4.9 & 51.1/48.8/\textbf{1.6}  & \textbf{87.7/83.1/1.6}  \\
TREC     & 24.7/18.6/6.6  & 27.1/18.2/\textbf{5.3} & \textbf{64.6/51.0}/6.1   \\ \midrule
CB       & 54.3/46.4/6.5 & 51.1/44.6/\textbf{4.1}  & \textbf{66.9/55.4}/4.8    \\
RTE      & 51.2/46.6/3.6 & 51.4/\textbf{46.9}/3.5 & \textbf{53.0}/44.4/\textbf{3.3}    \\
QNLI     & 49.4/49.2/\textbf{0.1}  & 49.5/\textbf{49.0}/0.3 & \textbf{52.8}/47.9/2.6    \\
MRPC     & \textbf{67.8/65.9/0.9} & 54.6/32.8/15.4 & 57.1/42.4/5.5    \\
QQP      & 37.0/36.8/\textbf{0.2} & 40.1/36.8/4.0 & \textbf{56.3/44.9}/4.1   \\
WiC      & 49.8/\textbf{49.7/0.1} & 49.8/48.6/0.5 & \textbf{51.1}/45.5/2.5  \\ \midrule
AVG      & 48.9/45.0/3.8  & 50.2/42.5/5.1 & \textbf{62.9/54.4/3.4}   \\
\bottomrule
\hline
\end{tabular}
}
\vspace{-8pt}
\caption{The performance of ICL-based methods when using BERT-large as the PLMs backbone. We report the average accuracy($\uparrow$)/worst-case accuracy($\uparrow$)/standard deviation($\downarrow$) and bold fonts indicate the best results.}
\label{table:bert-vs-roberta}
\end{table}

\section{Related Work}
\paragraph{Few-Shot Learning with PLMs.}
The GPT series~\cite{gpt18,Radford2019,gpt3} raise the attention of prompt-based learning. 
\citet{gpt3} propose the in-context learning and show the ability of PLMs to perform few-shot learning without any fine-tuning. 
In line with this work, \citet{calibrate} point out the bias issue in prompt-based PLMs and design contextual calibration method.
\citet{autoprompt} optimize prompt engineering with automatic prompt search.
\citet{lu-etal-2022-fantastically} present a generation-based probing method to decide ordering of prompts. 
\citet{rubin-etal-2022-learning} introduce retrieval modules to search prompts for  improving the quality of demonstrations. 
Our approach further exploits the potential of in-context learning by retrieving similar training examples to augment or correct the original decision boundary provided by PLMs.

More recently, substantial efforts have been made with optimizing the prompt format~\cite{le-scao-rush-2021-many,lester-etal-2021-power}. 
Several studies replace the manual prompts and verbalizers with continuous prompt embeddings~\cite{li-liang-2021-prefix,qin-eisner-2021-learning}, adapter layers~\cite{karimi-mahabadi-etal-2022-prompt}, and automatic generated ones~\cite{gao-etal-2021-making}.
Our proposed method takes a step forward, aiming to reduce the performance gap between in-context learning and existing fine-tuning methods. 

\begin{table}[t]
\centering
\small
\scalebox{0.95}{
\begin{tabular}{l|cc|c}
\hline    
\toprule
\multicolumn{1}{c|}{\multirow{2}{*}{Dataset}} & \multicolumn{2}{c|}{ICL} & KNN-C - ICL\\
\multicolumn{1}{c|}{}       &  Train & Test & on Test \\ \midrule
SST-2    & 90.2/81.2/5.3  & 89.9/88.3/1.0 & 2.7/1.8/-0.2   \\
SST-5    & 43.3/31.2/7.0 & 43.6/37.3/4.8 & 4.9/4.2/-2.3    \\
CR       & 87.2/78.1/3.4  & 88.0/83.1/2.5 & 2.5/2.9/-1.2    \\
MR       & 89.8/81.2/6.2  & 86.1/83.9/1.2  & 2.9/1.9/0  \\
SUBJ     & 52.8/43.8/5.3 & 51.2/48.5/2.9  & \textbf{39.3/37.4/-1.4} \\
TREC     & 47.2/37.5/7.6  & 52.7/36.8/10.3 & \textbf{24.6/27.4/-2.4}  \\ \midrule
CB       & 67.4/47.5/9.0 & 77.4/57.1/10.3 & 4.7/12.5/-5.0   \\
RTE      & 55.8/46.9/6.9 & 58.8/53.8/3.6 & 3.0/-1.4/-0.7    \\
QNLI     & 51.5/43.8/5.4 & 52.0/50.3/1.2 & 2.2/0.1/1.0   \\
MRPC     & 50.2/46.9/2.2 & 54.0/32.1/17.4 & \textbf{8.5/10.6/-10.1}   \\
QQP      & 52.2/34.4/7.9 & 42.5/36.8/6.7 & \textbf{16.2/10.6/-2.4}  \\
WiC      &  53.3/40.6/7.7 & 51.0/50.0/1.6 & 2.1/-3.8/1.5 \\
\bottomrule
\hline
\end{tabular}
}
\vspace{-8pt}
\caption{The performance of ICL on training and test data. We report the average accuracy($\uparrow$)/worst-case accuracy($\uparrow$)/standard deviation($\downarrow$) and highlight the 2-best improvements (i.e., KNN-C - ICL) among all tasks.}
\label{table:icl-errors}
\end{table}

\paragraph{Retrieval-Augmented Methods.}
Our proposed framework is closely related to the retrieval-augmented methods.
Recently, these approaches enhance the pre-trained models with a retrieval component and have achieved promising results in a variety of natural language processing tasks, including language modeling~\cite{pmlr-v119-guu20a,pmlr-v162-borgeaud22a}, machine translation~\cite{knnmt,zheng-etal-2021-adaptive,zheng-etal-2021-non-parametric,Wang2022NonParametricOL,Du2022NonParametricDA} and open-domain question answering~\cite{lewis20,izacard-grave-2021-leveraging}.
Recent simultaneous work~\cite{knn-prompt} propose a retrieval model to incorporates additional unsupervised data for zero-shot inference.
Different from them, we design a novel retrieval-augmented method for in-context learning, and augment PLMs with $k$NN retrieval constructed by few-shot support samples.



\section{Conclusion}
In this paper, we present a simple and effective nearest-neighbor calibration framework to improve the performance of in-context learning on few-shot text classification tasks. 
This approach directly augments PLMs with additional $k$NN classifier based on current few-shot support instances, where the adaptive neighbor selection and feature regularization modules are introduced to reduce the $k$NN retrieval noises.
Experimental results on various NLP tasks indicate that our method achieves significant improvements over in-context learning and is even comparable with the-state-of-art fine-tuning methods in single-sentence tasks. 
In the future, we would like to combine our method with bigger PLMs and further investigate the potential reason causing the performance gap between our method and current fine-tuning methods.
Another interesting direction is to explore our method on larger training datasets, rather than the few-shot adaptation setting. 



\bibliography{anthology,custom}

\begin{thebibliography}{42}
\expandafter\ifx\csname natexlab\endcsname\relax\def\natexlab#1{#1}\fi

\bibitem[{Agarap(2018)}]{relu}
Abien~Fred Agarap. 2018.
\newblock \href {http://arxiv.org/abs/1803.08375} {Deep learning using
  rectified linear units (relu)}.
\newblock \emph{CoRR}, abs/1803.08375.

\bibitem[{Ben~Zaken et~al.(2022)Ben~Zaken, Goldberg, and
  Ravfogel}]{ben-zaken-etal-2022-bitfit}
Elad Ben~Zaken, Yoav Goldberg, and Shauli Ravfogel. 2022.
\newblock \href {https://doi.org/10.18653/v1/2022.acl-short.1} {{B}it{F}it:
  Simple parameter-efficient fine-tuning for transformer-based masked
  language-models}.
\newblock In \emph{Proceedings of the 60th Annual Meeting of the Association
  for Computational Linguistics (Volume 2: Short Papers)}, pages 1--9, Dublin,
  Ireland. Association for Computational Linguistics.

\bibitem[{Borgeaud et~al.(2022)Borgeaud, Mensch, Hoffmann, Cai, Rutherford,
  Millican, Van Den~Driessche, Lespiau, Damoc, Clark, De~Las~Casas, Guy,
  Menick, Ring, Hennigan, Huang, Maggiore, Jones, Cassirer, Brock, Paganini,
  Irving, Vinyals, Osindero, Simonyan, Rae, Elsen, and
  Sifre}]{pmlr-v162-borgeaud22a}
Sebastian Borgeaud, Arthur Mensch, Jordan Hoffmann, Trevor Cai, Eliza
  Rutherford, Katie Millican, George~Bm Van Den~Driessche, Jean-Baptiste
  Lespiau, Bogdan Damoc, Aidan Clark, Diego De~Las~Casas, Aurelia Guy, Jacob
  Menick, Roman Ring, Tom Hennigan, Saffron Huang, Loren Maggiore, Chris Jones,
  Albin Cassirer, Andy Brock, Michela Paganini, Geoffrey Irving, Oriol Vinyals,
  Simon Osindero, Karen Simonyan, Jack Rae, Erich Elsen, and Laurent Sifre.
  2022.
\newblock \href {https://proceedings.mlr.press/v162/borgeaud22a.html}
  {Improving language models by retrieving from trillions of tokens}.
\newblock In \emph{Proceedings of the 39th International Conference on Machine
  Learning}, volume 162 of \emph{Proceedings of Machine Learning Research},
  pages 2206--2240. PMLR.

\bibitem[{Brown et~al.(2020)Brown, Mann, Ryder, Subbiah, Kaplan, Dhariwal,
  Neelakantan, Shyam, Sastry, Askell, Agarwal, Herbert-Voss, Krueger, Henighan,
  Child, Ramesh, Ziegler, Wu, Winter, Hesse, Chen, Sigler, Litwin, Gray, Chess,
  Clark, Berner, McCandlish, Radford, Sutskever, and Amodei}]{gpt3}
Tom Brown, Benjamin Mann, Nick Ryder, Melanie Subbiah, Jared~D Kaplan, Prafulla
  Dhariwal, Arvind Neelakantan, Pranav Shyam, Girish Sastry, Amanda Askell,
  Sandhini Agarwal, Ariel Herbert-Voss, Gretchen Krueger, Tom Henighan, Rewon
  Child, Aditya Ramesh, Daniel Ziegler, Jeffrey Wu, Clemens Winter, Chris
  Hesse, Mark Chen, Eric Sigler, Mateusz Litwin, Scott Gray, Benjamin Chess,
  Jack Clark, Christopher Berner, Sam McCandlish, Alec Radford, Ilya Sutskever,
  and Dario Amodei. 2020.
\newblock \href
  {https://proceedings.neurips.cc/paper/2020/file/1457c0d6bfcb4967418bfb8ac142f64a-Paper.pdf}
  {Language models are few-shot learners}.
\newblock In \emph{Advances in Neural Information Processing Systems},
  volume~33, pages 1877--1901. Curran Associates, Inc.

\bibitem[{De~Marneffe et~al.(2019)De~Marneffe, Simons, and
  Tonhauser}]{de2019commitmentbank}
Marie-Catherine De~Marneffe, Mandy Simons, and Judith Tonhauser. 2019.
\newblock The commitmentbank: Investigating projection in naturally occurring
  discourse.
\newblock In \emph{proceedings of Sinn und Bedeutung}, pages 107--124.

\bibitem[{Devlin et~al.(2019)Devlin, Chang, Lee, and Toutanova}]{bert}
Jacob Devlin, Ming-Wei Chang, Kenton Lee, and Kristina Toutanova. 2019.
\newblock \href {https://aclanthology.org/N19-1423} {{BERT}: Pre-training of
  deep bidirectional transformers for language understanding}.
\newblock In \emph{Proceedings of the 2019 Conference of the North {A}merican
  Chapter of the Association for Computational Linguistics: Human Language
  Technologies, Volume 1 (Long and Short Papers)}, pages 4171--4186,
  Minneapolis, Minnesota. Association for Computational Linguistics.

\bibitem[{Dolan and Brockett(2005)}]{mrpc}
William~B. Dolan and Chris Brockett. 2005.
\newblock \href {https://aclanthology.org/I05-5002} {Automatically constructing
  a corpus of sentential paraphrases}.
\newblock In \emph{Proceedings of the Third International Workshop on
  Paraphrasing ({IWP}2005)}.

\bibitem[{Du et~al.(2022)Du, Wang, Zhang, Chen, Xu, Xie, and
  Chen}]{Du2022NonParametricDA}
Yichao Du, Weizhi Wang, Zhirui Zhang, Boxing Chen, Tong Xu, Jun Xie, and Enhong
  Chen. 2022.
\newblock Non-parametric domain adaptation for end-to-end speech translation.
\newblock In \emph{EMNLP}.

\bibitem[{Gao et~al.(2021)Gao, Fisch, and Chen}]{gao-etal-2021-making}
Tianyu Gao, Adam Fisch, and Danqi Chen. 2021.
\newblock \href {https://doi.org/10.18653/v1/2021.acl-long.295} {Making
  pre-trained language models better few-shot learners}.
\newblock In \emph{Proceedings of the 59th Annual Meeting of the Association
  for Computational Linguistics and the 11th International Joint Conference on
  Natural Language Processing (Volume 1: Long Papers)}, pages 3816--3830,
  Online. Association for Computational Linguistics.

\bibitem[{Guu et~al.(2020)Guu, Lee, Tung, Pasupat, and
  Chang}]{pmlr-v119-guu20a}
Kelvin Guu, Kenton Lee, Zora Tung, Panupong Pasupat, and Mingwei Chang. 2020.
\newblock \href {https://proceedings.mlr.press/v119/guu20a.html} {Retrieval
  augmented language model pre-training}.
\newblock In \emph{Proceedings of the 37th International Conference on Machine
  Learning}, volume 119 of \emph{Proceedings of Machine Learning Research},
  pages 3929--3938. PMLR.

\bibitem[{Hu and Liu(2004)}]{hu2004mining}
Minqing Hu and Bing Liu. 2004.
\newblock Mining and summarizing customer reviews.
\newblock In \emph{Proceedings of the tenth ACM SIGKDD international conference
  on Knowledge discovery and data mining}, pages 168--177.

\bibitem[{Izacard and Grave(2021)}]{izacard-grave-2021-leveraging}
Gautier Izacard and Edouard Grave. 2021.
\newblock \href {https://doi.org/10.18653/v1/2021.eacl-main.74} {Leveraging
  passage retrieval with generative models for open domain question answering}.
\newblock In \emph{Proceedings of the 16th Conference of the European Chapter
  of the Association for Computational Linguistics: Main Volume}, pages
  874--880, Online. Association for Computational Linguistics.

\bibitem[{Karimi~Mahabadi et~al.(2022)Karimi~Mahabadi, Zettlemoyer, Henderson,
  Mathias, Saeidi, Stoyanov, and Yazdani}]{karimi-mahabadi-etal-2022-prompt}
Rabeeh Karimi~Mahabadi, Luke Zettlemoyer, James Henderson, Lambert Mathias,
  Marzieh Saeidi, Veselin Stoyanov, and Majid Yazdani. 2022.
\newblock \href {https://doi.org/10.18653/v1/2022.acl-long.254} {Prompt-free
  and efficient few-shot learning with language models}.
\newblock In \emph{Proceedings of the 60th Annual Meeting of the Association
  for Computational Linguistics (Volume 1: Long Papers)}, pages 3638--3652,
  Dublin, Ireland. Association for Computational Linguistics.

\bibitem[{Khandelwal et~al.(2020)Khandelwal, Fan, Jurafsky, Zettlemoyer, and
  Lewis}]{knnmt}
Urvashi Khandelwal, Angela Fan, Dan Jurafsky, Luke Zettlemoyer, and Mike Lewis.
  2020.
\newblock \href {http://arxiv.org/abs/2010.00710} {Nearest neighbor machine
  translation}.
\newblock \emph{CoRR}, abs/2010.00710.

\bibitem[{Kingma and Ba(2015)}]{adam}
Diederik~P. Kingma and Jimmy Ba. 2015.
\newblock \href {http://arxiv.org/abs/1412.6980} {Adam: {A} method for
  stochastic optimization}.
\newblock In \emph{3rd International Conference on Learning Representations,
  {ICLR} 2015, San Diego, CA, USA, May 7-9, 2015, Conference Track
  Proceedings}.

\bibitem[{Le~Scao and Rush(2021)}]{le-scao-rush-2021-many}
Teven Le~Scao and Alexander Rush. 2021.
\newblock \href {https://doi.org/10.18653/v1/2021.naacl-main.208} {How many
  data points is a prompt worth?}
\newblock In \emph{Proceedings of the 2021 Conference of the North American
  Chapter of the Association for Computational Linguistics: Human Language
  Technologies}, pages 2627--2636, Online. Association for Computational
  Linguistics.

\bibitem[{Lester et~al.(2021)Lester, Al-Rfou, and
  Constant}]{lester-etal-2021-power}
Brian Lester, Rami Al-Rfou, and Noah Constant. 2021.
\newblock \href {https://doi.org/10.18653/v1/2021.emnlp-main.243} {The power of
  scale for parameter-efficient prompt tuning}.
\newblock In \emph{Proceedings of the 2021 Conference on Empirical Methods in
  Natural Language Processing}, pages 3045--3059, Online and Punta Cana,
  Dominican Republic. Association for Computational Linguistics.

\bibitem[{Lewis et~al.(2020)Lewis, Perez, Piktus, Petroni, Karpukhin, Goyal,
  K\"{u}ttler, Lewis, Yih, Rockt\"{a}schel, Riedel, and Kiela}]{lewis20}
Patrick Lewis, Ethan Perez, Aleksandra Piktus, Fabio Petroni, Vladimir
  Karpukhin, Naman Goyal, Heinrich K\"{u}ttler, Mike Lewis, Wen-tau Yih, Tim
  Rockt\"{a}schel, Sebastian Riedel, and Douwe Kiela. 2020.
\newblock \href
  {https://proceedings.neurips.cc/paper/2020/file/6b493230205f780e1bc26945df7481e5-Paper.pdf}
  {Retrieval-augmented generation for knowledge-intensive nlp tasks}.
\newblock In \emph{Advances in Neural Information Processing Systems},
  volume~33, pages 9459--9474. Curran Associates, Inc.

\bibitem[{Li and Liang(2021)}]{li-liang-2021-prefix}
Xiang~Lisa Li and Percy Liang. 2021.
\newblock \href {https://doi.org/10.18653/v1/2021.acl-long.353} {Prefix-tuning:
  Optimizing continuous prompts for generation}.
\newblock In \emph{Proceedings of the 59th Annual Meeting of the Association
  for Computational Linguistics and the 11th International Joint Conference on
  Natural Language Processing (Volume 1: Long Papers)}, pages 4582--4597,
  Online. Association for Computational Linguistics.

\bibitem[{Liu et~al.(2019)Liu, Ott, Goyal, Du, Joshi, Chen, Levy, Lewis,
  Zettlemoyer, and Stoyanov}]{roberta}
Yinhan Liu, Myle Ott, Naman Goyal, Jingfei Du, Mandar Joshi, Danqi Chen, Omer
  Levy, Mike Lewis, Luke Zettlemoyer, and Veselin Stoyanov. 2019.
\newblock Roberta: A robustly optimized bert pretraining approach.
\newblock \emph{ArXiv}, abs/1907.11692.

\bibitem[{Logan~IV et~al.(2022)Logan~IV, Balazevic, Wallace, Petroni, Singh,
  and Riedel}]{logan-iv-etal-2022-cutting}
Robert Logan~IV, Ivana Balazevic, Eric Wallace, Fabio Petroni, Sameer Singh,
  and Sebastian Riedel. 2022.
\newblock \href {https://doi.org/10.18653/v1/2022.findings-acl.222} {Cutting
  down on prompts and parameters: Simple few-shot learning with language
  models}.
\newblock In \emph{Findings of the Association for Computational Linguistics:
  ACL 2022}, pages 2824--2835, Dublin, Ireland. Association for Computational
  Linguistics.

\bibitem[{Lu et~al.(2022)Lu, Bartolo, Moore, Riedel, and
  Stenetorp}]{lu-etal-2022-fantastically}
Yao Lu, Max Bartolo, Alastair Moore, Sebastian Riedel, and Pontus Stenetorp.
  2022.
\newblock \href {https://doi.org/10.18653/v1/2022.acl-long.556} {Fantastically
  ordered prompts and where to find them: Overcoming few-shot prompt order
  sensitivity}.
\newblock In \emph{Proceedings of the 60th Annual Meeting of the Association
  for Computational Linguistics (Volume 1: Long Papers)}, pages 8086--8098,
  Dublin, Ireland. Association for Computational Linguistics.

\bibitem[{Pang and Lee(2004)}]{subj}
Bo~Pang and Lillian Lee. 2004.
\newblock \href {https://doi.org/10.3115/1218955.1218990} {A sentimental
  education: Sentiment analysis using subjectivity summarization based on
  minimum cuts}.
\newblock In \emph{Proceedings of the 42nd Annual Meeting of the Association
  for Computational Linguistics ({ACL}-04)}, pages 271--278, Barcelona, Spain.

\bibitem[{Pang and Lee(2005)}]{mr}
Bo~Pang and Lillian Lee. 2005.
\newblock \href {https://doi.org/10.3115/1219840.1219855} {Seeing stars:
  Exploiting class relationships for sentiment categorization with respect to
  rating scales}.
\newblock In \emph{Proceedings of the 43rd Annual Meeting of the Association
  for Computational Linguistics ({ACL}{'}05)}, pages 115--124, Ann Arbor,
  Michigan. Association for Computational Linguistics.

\bibitem[{Perez et~al.(2021)Perez, Kiela, and Cho}]{perez2021true}
Ethan Perez, Douwe Kiela, and Kyunghyun Cho. 2021.
\newblock True few-shot learning with language models.
\newblock In \emph{Advances in Neural Information Processing Systems}.

\bibitem[{Pilehvar and Camacho-Collados(2019)}]{wic}
Mohammad~Taher Pilehvar and Jose Camacho-Collados. 2019.
\newblock \href {https://aclanthology.org/N19-1128} {{W}i{C}: the
  word-in-context dataset for evaluating context-sensitive meaning
  representations}.
\newblock In \emph{Proceedings of the 2019 Conference of the North {A}merican
  Chapter of the Association for Computational Linguistics: Human Language
  Technologies, Volume 1 (Long and Short Papers)}, pages 1267--1273,
  Minneapolis, Minnesota. Association for Computational Linguistics.

\bibitem[{Qin and Eisner(2021)}]{qin-eisner-2021-learning}
Guanghui Qin and Jason Eisner. 2021.
\newblock \href {https://doi.org/10.18653/v1/2021.naacl-main.410} {Learning how
  to ask: Querying {LM}s with mixtures of soft prompts}.
\newblock In \emph{Proceedings of the 2021 Conference of the North American
  Chapter of the Association for Computational Linguistics: Human Language
  Technologies}, pages 5203--5212, Online. Association for Computational
  Linguistics.

\bibitem[{Radford et~al.(2018)Radford, Narasimhan, Salimans, Sutskever
  et~al.}]{gpt18}
Alec Radford, Karthik Narasimhan, Tim Salimans, Ilya Sutskever, et~al. 2018.
\newblock Improving language understanding by generative pre-training.

\bibitem[{Radford et~al.(2019)Radford, Wu, Child, Luan, Amodei, Sutskever
  et~al.}]{Radford2019}
Alec Radford, Jeffrey Wu, Rewon Child, David Luan, Dario Amodei, Ilya
  Sutskever, et~al. 2019.
\newblock Language models are unsupervised multitask learners.
\newblock \emph{OpenAI blog}.

\bibitem[{Rajpurkar et~al.(2016)Rajpurkar, Zhang, Lopyrev, and Liang}]{qnli}
Pranav Rajpurkar, Jian Zhang, Konstantin Lopyrev, and Percy Liang. 2016.
\newblock \href {https://doi.org/10.18653/v1/D16-1264} {{SQ}u{AD}: 100,000+
  questions for machine comprehension of text}.
\newblock In \emph{Proceedings of the 2016 Conference on Empirical Methods in
  Natural Language Processing}, pages 2383--2392, Austin, Texas. Association
  for Computational Linguistics.

\bibitem[{Rubin et~al.(2022)Rubin, Herzig, and
  Berant}]{rubin-etal-2022-learning}
Ohad Rubin, Jonathan Herzig, and Jonathan Berant. 2022.
\newblock \href {https://doi.org/10.18653/v1/2022.naacl-main.191} {Learning to
  retrieve prompts for in-context learning}.
\newblock In \emph{Proceedings of the 2022 Conference of the North American
  Chapter of the Association for Computational Linguistics: Human Language
  Technologies}, pages 2655--2671, Seattle, United States. Association for
  Computational Linguistics.

\bibitem[{Schick and Sch{\"u}tze(2021)}]{pet}
Timo Schick and Hinrich Sch{\"u}tze. 2021.
\newblock \href {https://aclanthology.org/2021.eacl-main.20} {Exploiting
  cloze-questions for few-shot text classification and natural language
  inference}.
\newblock In \emph{Proceedings of the 16th Conference of the European Chapter
  of the Association for Computational Linguistics: Main Volume}, pages
  255--269, Online. Association for Computational Linguistics.

\bibitem[{Shi et~al.(2022)Shi, Michael, Gururangan, and
  Zettlemoyer}]{knn-prompt}
Weijia Shi, Julian Michael, Suchin Gururangan, and Luke Zettlemoyer. 2022.
\newblock \href {http://arxiv.org/abs/2205.13792} {Nearest neighbor zero-shot
  inference}.
\newblock \emph{CoRR}, abs/2205.13792.

\bibitem[{Shin et~al.(2020)Shin, Razeghi, IV, Wallace, and Singh}]{autoprompt}
Taylor Shin, Yasaman Razeghi, Robert L.~Logan IV, Eric Wallace, and Sameer
  Singh. 2020.
\newblock {AutoPrompt}: Eliciting knowledge from language models with
  automatically generated prompts.
\newblock In \emph{Empirical Methods in Natural Language Processing (EMNLP)}.

\bibitem[{Socher et~al.(2013)Socher, Perelygin, Wu, Chuang, Manning, Ng, and
  Potts}]{sst}
Richard Socher, Alex Perelygin, Jean Wu, Jason Chuang, Christopher~D. Manning,
  Andrew Ng, and Christopher Potts. 2013.
\newblock \href {https://aclanthology.org/D13-1170} {Recursive deep models for
  semantic compositionality over a sentiment treebank}.
\newblock In \emph{Proceedings of the 2013 Conference on Empirical Methods in
  Natural Language Processing}, pages 1631--1642, Seattle, Washington, USA.
  Association for Computational Linguistics.

\bibitem[{Taylor(1953)}]{taylor}
Taylor. 1953.
\newblock {cloze procedure: A new tool for measuring readability}.
\newblock \emph{Journalism quarterly}.

\bibitem[{Voorhees and Tice(2000)}]{trec}
Ellen~M Voorhees and Dawn~M Tice. 2000.
\newblock Building a question answering test collection.
\newblock In \emph{Proceedings of the 23rd annual international ACM SIGIR
  conference on Research and development in information retrieval}, pages
  200--207.

\bibitem[{Wang et~al.(2019)Wang, Pruksachatkun, Nangia, Singh, Michael, Hill,
  Levy, and Bowman}]{rte}
Alex Wang, Yada Pruksachatkun, Nikita Nangia, Amanpreet Singh, Julian Michael,
  Felix Hill, Omer Levy, and Samuel Bowman. 2019.
\newblock \href
  {https://proceedings.neurips.cc/paper/2019/file/4496bf24afe7fab6f046bf4923da8de6-Paper.pdf}
  {Superglue: A stickier benchmark for general-purpose language understanding
  systems}.
\newblock In \emph{Advances in Neural Information Processing Systems},
  volume~32. Curran Associates, Inc.

\bibitem[{Wang et~al.(2022)Wang, Wei, Zhang, Huang, Xie, Luo, and
  Chen}]{Wang2022NonParametricOL}
Dongqi Wang, Hao-Ran Wei, Zhirui Zhang, Shujian Huang, Jun Xie, Weihua Luo, and
  Jiajun Chen. 2022.
\newblock Non-parametric online learning from human feedback for neural machine
  translation.
\newblock In \emph{AAAI}.

\bibitem[{Zhao et~al.(2021)Zhao, Wallace, Feng, Klein, and Singh}]{calibrate}
Zihao Zhao, Eric Wallace, Shi Feng, Dan Klein, and Sameer Singh. 2021.
\newblock \href {http://proceedings.mlr.press/v139/zhao21c.html} {Calibrate
  before use: Improving few-shot performance of language models}.
\newblock In \emph{Proceedings of the 38th International Conference on Machine
  Learning, {ICML} 2021, 18-24 July 2021, Virtual Event}, volume 139 of
  \emph{Proceedings of Machine Learning Research}, pages 12697--12706. {PMLR}.

\bibitem[{Zheng et~al.(2021{\natexlab{a}})Zheng, Zhang, Guo, Huang, Chen, Luo,
  and Chen}]{zheng-etal-2021-adaptive}
Xin Zheng, Zhirui Zhang, Junliang Guo, Shujian Huang, Boxing Chen, Weihua Luo,
  and Jiajun Chen. 2021{\natexlab{a}}.
\newblock \href {https://doi.org/10.18653/v1/2021.acl-short.47} {Adaptive
  nearest neighbor machine translation}.
\newblock In \emph{Proceedings of the 59th Annual Meeting of the Association
  for Computational Linguistics and the 11th International Joint Conference on
  Natural Language Processing (Volume 2: Short Papers)}, pages 368--374,
  Online. Association for Computational Linguistics.

\bibitem[{Zheng et~al.(2021{\natexlab{b}})Zheng, Zhang, Huang, Chen, Xie, Luo,
  and Chen}]{zheng-etal-2021-non-parametric}
Xin Zheng, Zhirui Zhang, Shujian Huang, Boxing Chen, Jun Xie, Weihua Luo, and
  Jiajun Chen. 2021{\natexlab{b}}.
\newblock \href {https://doi.org/10.18653/v1/2021.findings-emnlp.358}
  {Non-parametric unsupervised domain adaptation for neural machine
  translation}.
\newblock In \emph{Findings of the Association for Computational Linguistics:
  EMNLP 2021}, pages 4234--4241, Punta Cana, Dominican Republic. Association
  for Computational Linguistics.

\end{thebibliography}
\bibliographystyle{acl_natbib}

\clearpage
\appendix

\section{Appendix}

\subsection{Comparisons with Fine-tuning Methods }
\label{sec:appendix-compare-ft}
Table~\ref{table:all-results} illustrates the performance of ICL-based and FT-based methods on all datasets.
In addition to FT, PET, and PERFECT, we include the results of other parameter-efficient tuning methods, including +Null-Prompt~\cite{logan-iv-etal-2022-cutting} and BitFit+mte~\cite{ben-zaken-etal-2022-bitfit}.
Among FT-based methods, PERFECT has obtained the best performance in 12 datasets on average.
Surprisingly, KNN-C achieves better or comparable results than PERFECT in five single-sentence tasks.
Another appealing capability of our method is to keep one model for all tasks in the Language-Model-as-a-Service setting and maintain knowledge for different tasks by external datastore. 
This way could further promote the landing of Language-Model-as-a-Service.

Our approach requires the training of ANS and FR modules, but the training cost is far less than FT-based methods.
We take the example of RoBERTa-large, which is used as the PLMs backbone in our paper: (1) the tunable parameters in our model are 32x32=1024 for the ANS module, and 1024x32+32=32800 for the FR module, thus KNN-C has 32800+1024=33824=0.03M tunable parameters in total; (2) fine-tuning the whole RoBERTa-large model (i.e., PET) or additional adapter layer (i.e., PERFECT) requires tuning 355M/3.3M parameters, respectively. 
In addition, KNN-C only needs one forward computation of the PLMs during the entire training process and the remaining computation is very small, thus the training time is far less than FT-based methods.
For the inference time, KNN-C is almost the same as ICL, as the datastore is very tiny and performing $k$NN retrieval is negligible.

\begin{table*}[t]
\small
\centering
\scalebox{0.9}{
\begin{tabular}{l|cccccc|c}
\hline  
\toprule
\multicolumn{1}{l|}{Methods} & \multicolumn{1}{c}{\textbf{SST-2}} & \multicolumn{1}{c}{\textbf{SST-5}} & \multicolumn{1}{c}{\textbf{CR}}   & \multicolumn{1}{c}{\textbf{MR}}   & \multicolumn{1}{c}{\textbf{SUBJ}} & \multicolumn{1}{c}{\textbf{TREC}} & \multicolumn{1}{|c}{\textbf{AVG}} \\ \midrule
\multicolumn{8}{c}{\textbf{FT-based Methods}}     \\ \midrule
FT   & 81.4/70.0/4.0  & 39.2/34.3/2.5  & 80.1/72.9/4.1  & 77.7/66.8/4.6  & \textbf{90.2/84.1/1.8}    & 87.6/75.8/3.7   & 76.0/67.3/3.4    \\
PET    & 89.7/81.0/2.4   & \textbf{45.9}/40.3/2.4  & 88.4/68.8/3.0   & 85.9/79.0/2.1   & 88.1/79.6/2.4    & 85.0/70.6/4.5   & 80.5/69.9/2.8  \\
+Null-Prompt & 89.8/84.1/1.7 & 45.7/\textbf{41.6/2.3}  & 89.9/87.2/1.1   & 84.9/76.2/3.2  & 81.8/73.5/4.0 & 84.7/81.8/1.6  & 79.5/74.1/2.3  \\ 
BitFit+mte  & 89.5/81.7/3.0  & 42.3/36.8/3.3 & \textbf{90.1/87.8/1.0}  & 85.6/80.5/1.9  & 89.1/82.4/2.4   & 90.4/85.0/1.4  & 81.2/75.7/2.2  \\ 
PERFECT   & \textbf{90.7/88.2/1.2}  & 42.7/35.1/2.9  & 90.0/85.5/1.4  & \textbf{86.3/81.4/1.6}  & 89.1/82.8/2.1  & \textbf{90.6/81.6/3.2}  & \textbf{81.6/75.8/2.1}  \\ \midrule
\multicolumn{8}{c}{\textbf{ICL-based Methods}}      \\ \midrule
PV-Zero$^{*}$   & 63.4/54.0/11.8 & 27.9/21.1/4.8   & 59.6/50.4/11.8  & 62.6/53.7/10.9                    & 55.5/50.2/7.6   & 33.9/25.6/8.8     & 50.5/42.5/9.3    \\
ICL & 89.9/88.3/1.0  & 43.6/37.3/4.8  & 88.0/83.1/2.5 & 86.1/83.9/1.2  & 51.2/48.5/2.9   &  52.7/36.8/10.3   & 68.6/63.0/3.8  \\ 
+ CC & 84.6/82.3/1.9  & 38.3/33.2/3.8 & 85.4/80.9/2.5 & 80.7/77.1/2.3  & 53.4/50.5/2.8   &  49.8/33.4/9.1   & 65.4/59.6/3.7  \\ 
KNN-C      & \textbf{92.6/90.1/0.8}  & \textbf{48.5/41.5/2.5}  &  \textbf{90.5/86.0/1.3}   & \textbf{89.0/85.8/1.2}  & \textbf{90.5/86.9/1.5} & \textbf{77.3/64.2/7.9}   & \textbf{81.4/75.7/2.5}  \\ \midrule
\multicolumn{1}{l|}{Methods} & \multicolumn{1}{c}{\textbf{CB}}    & \multicolumn{1}{c}{\textbf{RTE}}   & \multicolumn{1}{c}{\textbf{QNLI}} & \multicolumn{1}{c}{\textbf{MRPC}} & \multicolumn{1}{c}{\textbf{QQP}}  & \multicolumn{1}{c}{\textbf{WiC}}  & \multicolumn{1}{|c}{\textbf{AVG}} \\ \midrule 
\multicolumn{8}{c}{\textbf{FT-based Methods}}   \\ \midrule
FT  & 72.9/67.9/\textbf{2.5}  & 56.8/50.2/3.5 & 62.7/51.4/7.0   & \textbf{70.1/62.7/4.7} & 65.0/59.8/3.6   & 52.4/46.1/3.7   & 63.3/56.4/4.2  \\
PET     & 86.9/73.2/5.1   & 60.1/49.5/4.7   & 66.5/55.7/6.2  & 62.1/38.2/6.8 & 63.4/44.7/7.9 & 51.0/46.1/2.6   & 65.0/51.2/5.6 \\
+Null-Prompt & \textbf{91.0/87.5}/2.7  & \textbf{64.4/58.5/3.9} & 71.2/\textbf{66.5/2.6}  & 63.9/53.7/5.3 & 70.4/62.7/3.4  & 52.4/\textbf{48.4/1.8} & 68.9/\textbf{62.9/3.3} \\ 
BitFit+mte  & 89.6/82.1/4.3  & 61.3/53.8/5.2  & 70.6/51.9/5.9  & 68.5/57.4/5.1 & 69.4/63.0/3.9 & 52.9/47.8/2.7   & 68.7/59.3/4.5  \\ 
PERFECT  & 90.3/83.9/3.5  & 60.4/53.1/4.7 & \textbf{74.1}/60.3/4.6  & 67.8/54.7/5.7  & \textbf{71.2/64.2/3.5} & \textbf{53.8}/47.0/3.0 & \textbf{69.6}/60.5/4.2  \\ \midrule
\multicolumn{8}{c}{\textbf{ICL-based Methods}}       \\ \midrule
PV-Zero$^{*}$  & 64.3/58.9/\textbf{3.4}   & 56.7/\textbf{54.2}/1.8  & 50.1/49.4/\textbf{0.6}  & \textbf{67.5/66.4/0.7} & 36.8/36.8/\textbf{0.1}   & 51.4/48.8/2.9   & 54.5/\textbf{52.4/1.6}  \\
ICL & 77.4/57.1/10.3 & 58.8/53.8/3.6 & 52.0/50.3/1.2 & 54.0/32.1/17.4  & 42.5/36.8/6.7   & 51.0/\textbf{50.0}/1.6  & 55.9/46.7/6.8 \\
+ CC & 56.9/42.9/10.0 & 56.0/53.8/\textbf{1.6} & \textbf{54.4/52.8}/0.8 & 62.2/37.8/10.4  & 45.0/36.7/6.6   & 49.3/47.5/\textbf{1.0} & 54.0/45.2/5.1  \\ 
KNN-C       & \textbf{82.1/69.6}/5.3  & \textbf{61.8}/52.4/2.9 & 54.2/50.4/2.2    & 62.5/42.7/7.3 & \textbf{58.7/47.4}/4.3 & \textbf{53.1}/46.2/3.1  & \textbf{62.1}/51.4/4.2 \\ 
\bottomrule
\hline
\end{tabular}
}
\vspace{-8pt}
\caption{Performance of all methods on 6 single-sentence (top) and 6 sentence-pair (bottom) benchmarks. $*$: no training examples are used. We report the average accuracy($\uparrow$)/worst-case accuracy($\uparrow$)/standard deviation($\downarrow$). Bold fonts indicate the best results.}
\label{table:all-results}
\end{table*}

\end{document}